\documentclass[letterpaper]{article}
\usepackage[letterpaper]{geometry}

\usepackage[utf8]{inputenc}

\usepackage{amsmath}
\usepackage{graphicx}
\usepackage{url}
\usepackage{xcolor}
\usepackage{latexsym}

\usepackage{hyperref}
\definecolor{darkblue}{rgb}{0, 0, 0.5}
\hypersetup{colorlinks=true,citecolor=darkblue, linkcolor=darkblue, urlcolor=darkblue}
\usepackage{tabularx}
\usepackage{datetime}
\usepackage{arydshln}

\usepackage{setspace}

\usepackage[authoryear,round,longnamesfirst]{natbib}
\usepackage[ruled,vlined]{algorithm2e}
\usepackage[linguistics]{forest} 
\usepackage{synttree}
\usepackage{subcaption}
\usepackage{tikz-dependency}

\usepackage{gb4e}

\title{Evaluating Prompting Strategies for Grammatical Error Correction Based on Language Proficiency
}
\author{ 
Min Zeng$^{1*}$~~
Jiexin Kuang$^{1}$\thanks{*Equally contributed authors.}~~
Mengyang Qiu$^{2}$~~
Jayoung Song$^{3}$~~
Jungyeul Park$^{1}$\\
$^{1}$Department of Linguistics, The University of British Columbia, Canada\\
$^{2}$Department of Psychology, Trent University, Canada\\
$^{3}$Department of Asian Studies, Pennsylvania State University, USA \\
{\tt \{minzengz,jxkuang\}@student.ubc.ca}~~~~~
{\tt mengyangqiu@trentu.ca}~~~~~\\
{\tt jayoung.song@psu.edu}~~~~~
{\tt jungyeul@mail.ubc.ca}\\
}
\date{ 
~\\
Preprint version. To appear in \textit{LREC-COLING 2024}, short paper. \\
}

\begin{document}

\maketitle

\begin{abstract}
This paper proposes an analysis of prompting strategies for grammatical error correction (GEC) with selected large language models (LLM) {based on language proficiency}. GEC using generative LLMs has been known for overcorrection where results obtain higher recall measures than precision measures.
The writing examples of English language learners may be different from those of native speakers.
{Given that there is a significant differences in second language (L2) learners' error types by their proficiency levels, this paper attempts to reduce overcorrection by examining the interaction between LLM's performance and L2 language proficiency.}
Our method focuses on zero-shot and few-shot prompting and fine-tuning models for GEC for learners of English as a foreign language {based on the different proficiency}.
We investigate GEC results and find that overcorrection happens primarily in advanced language learners’ writing (proficiency C) rather than proficiency A (a beginner level) and proficiency B (an intermediate level). 
Fine-tuned LLMs, and even few-shot prompting with writing examples of English learners, actually tend to exhibit decreased recall measures. 
To make our claim concrete, we conduct a comprehensive examination of GEC outcomes and their evaluation results based on language proficiency.
\end{abstract}
\doublespacing

\section{Introduction}

Large language models (LLMs) like Generative Pre-trained Transformers (GPT) have emerged as a transformative force in natural language processing (NLP) and artificial intelligence. 
These models, boasting billions of parameters, have been trained on an extensive corpus of internet text, making them highly effective across a wide spectrum of language tasks, such as translation, summarization, and question answering, often achieving state-of-the-art results \citep{brown-etal-2020-nips}.

One such application of LLMs is Grammatical Error Correction (GEC). 
GEC is a challenging task in NLP that involves detecting and correcting grammatical mistakes in written text. 
LLMs like GPT have shown promising results in this domain, with their ability to generate fluent, grammatically correct text \citep[e.g.,][]{coyne2023analyzing,loem-etal-2023-exploring}.
However, despite their impressive performance, these models are not without limitations.
For example, LLMs have a tendency to overcorrect, leading to higher recall but lower precision measures \citep{fang-et-al-2023-chatgpt}.

Grammatical Error Correction has been a pivotal task in NLP, with numerous methodologies and systems being developed over the years to improve its performance.
Prior to the advent of LLMs, the most effective GEC systems have predominantly adopted one of two paradigms: sequence-to-sequence Neural Machine Translation (NMT)-based approaches and sequence tagging edit-based approaches.
The unique characteristic of GEC, notably the high overlap between the source and target sentences, has led to the development of edit-based approaches. 
These models employ a transformer-based architecture, akin to their NMT-based counterparts. 
However, instead of predicting entire sentences, they are trained to anticipate a sequence of editing operations, such as delete, append, and replace, significantly enhancing the speed of inference while preserving high performance \citep{omelianchuk-etal-2020-gector}.

The advent of LLMs has ushered in a new era for GEC. 
A notable example of is the work by \citet{rothe-etal-2021-simple}, where they leveraged the power of LLMs, specifically the mT5 model with up to 11 billion parameters.
Their work establishes a new set of baselines for GEC and simplifies the typical GEC training pipelines composed of multiple fine-tuning stages.

In addition to this fine-tuning approach, recent studies have begun to explore the potential of the prompt-based approach in the application of LLMs for GEC, which focuses more on the design of effective prompts that guide the model's generation of corrected sentences.
For example, \citet{loem-etal-2023-exploring} investigated the impact of task instructions and the number of examples on the performance of GPT-3 in GEC tasks. 
They found that instructive instructions significantly improved GPT-3's performance in both zero-shot and few-shot settings, and the performance became more consistent as it received more examples. 

Another area which should be taken into account is L2 learners' language proficiency levels. Considering that there is a significant relationship between learners’ language proficiency levels and types of errors they make \citep{yuksel-inan-fidan-2017}, having language proficiency as one of the variables in the model might enhance the performance of the model. To be specific, exploring the relationships between GEC using LLMs, especially, GPT, and language proficiency levels could reduce the notable limitation of LLMs, that it its tendency to overcorrection, leading to higher recall but lower precision measures \citep{fang-et-al-2023-chatgpt}.

Building upon these observations, this paper intends to explore the performance of LLMs in GEC {by examining the interaction between LLMs' performance and the language proficiency levels of the learners.}
We focus our exploration on how prompting strategies and fine-tuning impact GEC performance, with particular attention given to zero-shot and few-shot prompting. 
Our goal is to provide a comprehensive understanding of the strengths and limitations of LLMs in GEC, aiming to illuminate ways in which their performance can be optimized for {language} learners of different proficiency levels, {which has hardly been explored thoroughly.}

\section{Language Proficiency} \label{language-proficiency-section}

{For prompting GEC using GPTs, we use the Cambridge English Write \& Improve} (W\&I) corpus, which is manually annotated with CEFR proficiency levels, consisting of beginner level A, intermediate level B, and advanced level C {\citep{yannakoudakis-etal-2018-wi}. It was introduced at the \textit{Building Educational Applications 2019 Shared Task: Grammatical Error Correction} (BEA2019)  \citep{bryant-etal-2019-bea}}. The text data was from writings of L2 English learners. It has a propensity that sentences from data of higher proficiency are longer than lower proficiency: average tokens per sentence in training data sets A, B, and C are 17.538, 18.304, and 19.212, respectively. For {a characteristic example of proficiency A}, the case of \textit{in} in the ungrammatical sentence \eqref{proficiency-1a-sent} is corrected with \textit{In} in its counterpart correction \eqref{proficiency-1b-sent}. It showcases a typical replacement orthography error, to be more specific, a capitalization error. We can also observe that the sentence contains an error \textit{with}, {which is corrected with} \textit{that} (\texttt{R:PREP}). Although it is grammatically accurate to use \textit{agree with} as a transitive phrasal verb, {an object clause} of the verb in the example sentence is not grammatical. 
In this case, the error annotation scheme maintains the structure of the clause while replacing the preposition instead.

\begin{exe}
{
\ex \label{proficiency-1a-sent} *{\color{red}in} addition more and more scientists agree {\color{red}with} {\color{red}alien} really exist
\ex \label{proficiency-1b-sent} {\color{white}*}{\color{green}In} addition{\color{green},} more and more scientists agree {\color{green}that} {\color{green}aliens} really exist{\color{green}.}
}
\end{exe}

We analyze the error distribution in training data of different language proficiency levels, in which the distribution of errors in the data sets of proficiency levels B and C is similar: 
missing punctuation marks (\texttt{M:PUNCT}), replacement prepositions (\texttt{R:PREP}), and missing determinants (\texttt{M:DET}) are {the most apparent types of errors.} 
Additionally, proficiency A includes an extra error type, replacement orthography (\texttt{R:ORTH}), which is defined {for} case or whitespace errors. 
Table~\ref{top-error-type} shows the ratios of the most frequent error types {of training data in W\&I, which we investigate thoroughly in $\S$\ref{discussion-section}.}

\begin{table}[!ht]
\centering
\begin{center}
{
\footnotesize
\begin{tabular}{cc | cc | cc} \hline
\multicolumn{2}{c|}{Proficiency A} & \multicolumn{2}{c|}{Proficiency B} & \multicolumn{2}{c}{Proficiency C}\\ \hline 
M:PUNCT & 0.0933 & M:PUNCT & 0.1134 & M:PUNCT & 0.1183 \\  
R:ORTH & 0.0602 & R:PREP & 0.0589 & R:PREP & 0.0517 \\  
R:PREP & 0.0506 & M:DET & 0.0442 & M:DET & 0.0345 \\  
R:VERB:TENSE & 0.0455 & R:VERB & 0.0414 & R:VERB & 0.0323 \\  
R:VERB & 0.0419 & R:VERB:TENSE & 0.0393 & R:VERB:TENSE & 0.0273 \\  
\hline
    \end{tabular}
}
\end{center}  
\caption{{Most frequent errors and their ratio in W\&I}}
\label{top-error-type}
\end{table}

\section{{Experimental Results}} \label{results-section}

For experiments, we use the development data set {of W\&I} from BEA2019, which distinguishes language proficiency levels into A, B and C. 
We follow the experimental setting described in \citet{suzgun-etal-2022-prompt} for GPT-2 (\texttt{gpt2-xl}) inferences, and we also adapt it to GPT-3.5 (\texttt{text-davinci-003}).
Instead of using the test data set for the BEA2019 \citep{bryant-etal-2019-bea}, we use the development data set for evaluation to control proficiency levels. 
To evaluate the performance of language proficiency levels  A, B, and C, we report ERRANT results \citep{bryant-felice-briscoe:2017:ACL} as metrics that include true positive, false positive, false negative, precision, recall, and more importantly, F0.5 scores which emphasize precision than recall. 
Table~\ref{gpt2-xl-results} summarizes the prompting GEC results for different language models, including GPT-2, GPT-3.5, and fine-tuned GPT-2. 
We used default setting in \citet{suzgun-etal-2022-prompt} for {inference parameters}:

\begin{center}
{
{\footnotesize
\begin{tabular} {r l}  \hline
model & \texttt{gpt2-xl} \\
tokenizer & \texttt{gpt2-xl} \\
num\_examplars & 0-4 shots \\
max\_model\_token\_length & 256 if num\_examplars is 0 \\
& else 512 \\
delimiter left and right & \{ \} \\ 
\hline
\end{tabular}
}}
\end{center}  

\begin{table*}[ht]
    \centering
\resizebox{\textwidth}{!}
{
{\footnotesize
\begin{tabularx}{\textwidth}{c|  r|X} \hline
{1-shot}&ungrammatical     &  This is important thing. \\
{\color{white}A}&grammatical     & This is an important thing. \\\hdashline
{2-shot}&ungrammatical     &  Water is needed for alive. \\
{\color{white}A}&grammatical     & Water is necessary to live. \\\hdashline
{3-shot}&ungrammatical     &  And young people spend time more ther lifestile. \\
{\color{white}A}&grammatical     & And young people spend more time on their lifestyles. \\\hdashline
{4-shot}&ungrammatical     &  Both of these men have dealed with situations in an unconventional manner and the results are with everyone to see. \\
{\color{white}A}&grammatical     &  Both of these men have dealt with situations in an unconventional manner and the results are plain to see.\\\hline
\end{tabularx}
}}
    \caption{Prompt examples}
    \label{prompt-table}
\end{table*}

We used the {prompts} described in Table~\ref{prompt-table}, and the following setting for {fine-tuning parameters}:

\begin{center}
{\footnotesize
\begin{tabular} {r l} 
\hline
epochs & 5 \\
using masked language modeling & False \\
block size (train) & 128 \\
per\_device\_train\_batch\_size & 4 \\
save\_steps & 10000 \\
save\_total\_limit & 2 \\
\hline
\end{tabular}
}
\end{center}


\begin{table*}
    \centering
\resizebox{\textwidth}{!}{    
\begin{tabular}{ r r| cccccc | cccccc| cccccc | ccc ccc} \hline 
&&  \multicolumn{6}{c|}{A} & \multicolumn{6}{c|}{B} & \multicolumn{6}{c|}{C}  & \multicolumn{6}{c}{all}\\ 
&& TP & FP & FN &  Prec & Rec & F0.5 & TP & FP & FN & Prec & Rec & F0.5& TP & FP & FN & Prec & Rec & F0.5  & TP & FP & FN & Prec & Rec & F0.5\\ \hline

\textsc{GPT-2} 

& zero-shot & 70 & 3944 & 2878 & 0.0174 & 0.0237 & 0.0184 & 45 & 5204 & 2453 & 0.0086 & 0.018 & 0.0096 & 28 & 4860 & 1058 & 0.0057 & 0.0258 & 0.0068 & 143 & 14008 & 6389 & 0.0101 & 0.0219 & 0.0113 \\
& 1-shot & 86 & 3447 & 2862 & 0.0243 & 0.0292 & 0.0252 & 58 & 4240 & 2440 & 0.0135 & 0.0232 & 0.0147 & 28 & 3730 & 1058 & 0.0075 & 0.0258 & 0.0087 & 172 & 11417 & 6360 & 0.0148 & 0.0263 & 0.0163  \\ 
& 2-shot & 103 & 4175 & 2845 & 0.0241 & 0.0349 & 0.0257 & 69 & 5442 & 2429 & 0.0125 & 0.0276 & 0.0141 & 30 & 4905 & 1056 & 0.0061 & 0.0276 & 0.0072 & 202 & 14522 & 6330 & 0.0137 & 0.0309 & 0.0154 \\ 
& 3-shot & 140 & 4445 & 2808 & 0.0305 & 0.0475 & \textbf{0.0329} & 95 & 5710 & 2403 & 0.0164 & 0.038 & \textbf{0.0185} & 38 & 4979 & 1048 & 0.0076 & 0.035 & \textbf{0.009} & 273 & 15134 & 6259 & 0.0177 & 0.0418 & \textbf{0.02} \\ 
& 4-shot & 133 & 4347 & 2815 & 0.0297 & 0.0451 & 0.0319 & 84 & 5422 & 2414 & 0.0153 & 0.0336 & 0.0171 & 31 & 4790 & 1055 & 0.0064 & 0.0285 & 0.0076 & 248 & 14559 & 6284 & 0.0167 & 0.038 & 0.0189  \\ \hdashline

\textsc{GPT-3.5} 
& zero-shot & 1203 & 3770 & 1740 & 0.2419 & 0.4088 & 0.2634 & 940 & 4693 & 1556 & 0.1669 & 0.3766 & 0.1878  & 407 & 4183 & 677 & 0.0887 & 0.3755 & 0.1047 & 2550 & 12646 & 3973 & 0.1678 & 0.3909 & 0.1894 \\
& 1-shot & 1300 & 3086 & 1643 & 0.2964 & 0.4417 & 0.3173 & 1068 & 3562 & 1428 & 0.2307 & 0.4279 & 0.2541 & 472 & 3086 & 612 & 0.1327 & 0.4354 & 0.1541 & 2840 & 9734 & 3683 & 0.2259 & 0.4354 & 0.2499\\ 
& 2-shot & 1443 & 2983 & 1500 & 0.326 & 0.4903 & 0.3494 & 1116 & 3157 & 1380 & 0.2612 & 0.4471 & 0.2849 & 486 & 2592 & 598 & 0.1579 & 0.4483 & 0.1814 & 3045 & 8732 & 3478 & 0.2586 & 0.4668 & 0.2839\\ 
& 3-shot & 1477 & 2646 & 1466 & 0.3582 & 0.5019 & \textbf{0.38} & 1114 & 3164 & 1382 & 0.2604 & 0.4463 & 0.2841 & 479 & 2416 & 605 & 0.1655 & 0.4419 & 0.1891 & 3070 & 8226 & 3453 & 0.2718 & 0.4706 & 0.2969\\ 
& 4-shot & 1330 & 2328 & 1613 & 0.3636 & 0.4519 & 0.3784 & 1089 & 2424 & 1407 & 0.31 & 0.4363 & \textbf{0.329} & 457 & 1870 & 627 & 0.1964 & 0.4216 & \textbf{0.2199} & 2876 & 6622 & 3647 & 0.3028 & 0.4409 & \textbf{0.323}\\ \hdashline

\textsc{FT GPT-2} 
& zero-shot & 1118 & 1479 & 1830 & 0.4305 & 0.3792 & \textbf{0.4192} & 928 & 1203 & 1570 & 0.4355 & 0.3715 & \textbf{0.421} & 383 & 792 & 703 & 0.326 & 0.3527 & \textbf{0.331} & 2429 & 3474 & 4103 & 0.4115 & 0.3719 & \textbf{0.4029}\\
& 1-shot & 1127 & 1668 & 1821 & 0.4032 & 0.3823 & 0.3989 & 925 & 1325 & 1573 & 0.4111 & 0.3703 & 0.4022 & 382 & 913 & 704 & 0.295 & 0.3517 & 0.3048 & 2434 & 3906 & 4098 & 0.3839 & 0.3726 & 0.3816\\ 
& 2-shot & 1107 & 1700 & 1841 & 0.3944 & 0.3755 & 0.3904 & 937 & 1359 & 1561 & 0.4081 & 0.3751 & 0.401 & 383 & 919 & 703 & 0.2942 & 0.3527 & 0.3043 & 2427 & 3978 & 4105 & 0.3789 & 0.3716 & 0.3774\\ 
& 3-shot & 1073 & 1860 & 1875 & 0.3658 & 0.364 & 0.3655 & 874 & 1596 & 1624 & 0.3538 & 0.3499 & 0.353 & 381 & 1168 & 705 & 0.246 & 0.3508 & 0.2616 & 2328 & 4624 & 4204 & 0.3349 & 0.3564 & 0.339\\ 
& 4-shot & 1032 & 1911 & 1916 & 0.3507 & 0.3501 & 0.3505 & 818 & 1815 & 1680 & 0.3107 & 0.3275 & 0.3139 & 359 & 1310 & 727 & 0.2151 & 0.3306 & 0.2313 & 2209 & 5036 & 4323 & 0.3049 & 0.3382 & 0.311 \\ \hline


SOTA & \textsc{gector} 
& 1046 & 632 & 2054 & 0.6234 & 0.3374 & 0.533 & 785 & 458 & 1836 & 0.6315 & 0.2995 & 0.5169 & 315 & 208 & 845 & 0.6023 & 0.2716 & 0.4843 & 2146 & 1298 & 4735 & 0.6231 & 0.3119 & 0.5194 \\ 
& \textsc{t5} 
& 1338 & 741 & 1762 & 0.6436 & 0.4316 & 0.586 & 1018 & 620 & 1603 & 0.6215 & 0.3884 & 0.5549 & 377 & 351 & 783 & 0.5179 & 0.325 & 0.4629 & 2733 & 1712 & 4148 & 0.6148 & 0.3972 & 0.5541 \\ \hline 
\end{tabular}
}\caption{Prompting results using GPT-2 (\texttt{gpt2-xl} and \textsc{ft} = fine-tuned), GPT-3.5 (\texttt{text-davinci-003}) and SOTA results by models of \textsc{gector} \citep{omelianchuk-etal-2020-gector} and \textsc{t5} \citep{rothe-etal-2021-simple}.} \label{gpt2-xl-results}
\end{table*}

When evaluating the efficacy of few-shot strategies on GPT-2 and GPT-3.5, it is evident that the few-shot prompting method exhibits better performance compared to the zero-shot prompting method. For instance, in the \texttt{all} data set which combines corpus of three language proficiency levels, we observe that the 4-shot F0.5 scores for GPT-2 and GPT-3.5 are 0.0495 and 0.323 respectively, which are higher than the zero-shot F0.5 scores for GPT-2 and GPT-3.5. It is also noticeable that the 4-shot approach consistently yields higher F0.5 scores in comparison to the zero-shot approach. 
However, this trend is not observed for the fine-tuned GPT-2 model on different language proficiency levels. For example, in the \texttt{all}  data set, the F0.5 score for the 4-shot approach is lower than the F0.5 score for the zero-shot approach. Therefore, based on our experimental findings, it is feasible to conclude that few-shot techniques may not have a significant impact on fine-tuned GPT-2 models.

In addition, GPT-2 exhibits a large decreasing rate of recall as the language proficiency levels increase from A to C. Specifically, there is a notable increase in the dropping rate of precision from 50.57\% (0.0174 in A versus 0.0086 in B) to 33.72\% (0.0086 in B versus 0.0057 in C). However, the fine-tuned GPT-2 shows a better trend for the precision rate. From proficiency level A to proficiency level C, the precision score increases from 0.4305 in A to 0.4355 in B (+1.16\%) and then drops to 0.326 (-25.14\%) in C. It indicates the fine-tuned model is more robust for different proficiency level data sets.

\section{Analysis and Discussion}\label{discussion-section}

{Unless specified otherwise, our analysis and discussion are based on results of the fine-tuned \texttt{gpt2-xl} using zero-shot which we achieve the best results.}

\paragraph{Label-by-label evaluation approach}
{We implement a label-by-label evaluation method. 
As \citet{bryant-felice-briscoe:2017:ACL} suggested, we provide edit operation-based and POS-based errors as well as \textit{detailed breakdown} composed errors (\textsc{m|r|u} with POS) to investigate further the relationship between GEC and different proficiency levels.
{For example, Table~\ref{label-by-label} shows different types of error evaluation results.}
When comparing correcting missing operation errors with all errors, it has higher F0.5 scores where it suggests that GEC using GPT performs better in the specific missing error {regardless of language proficiency.
\texttt{M:PUNCT} (missing punctuation marks) is the most frequent error among all error types in three language proficiency, which outperforms the \textsc{entire} results for all proficiency levels. This reflects the general characteristics of the performance of GEC using GPT.}
{\texttt{R:VERB} (replacing verbs) consistently performs poorly compared to the entire results, and this has the same tendency for all \textsc{r} edit errors where the proficiency C achieves especially lower results.
We observed that GEC using GPT contradicts to the problem of over-correction for lower proficiency levels because of the much higher numbers of FN in A and B.}

\begin{table}[!ht]
    \centering
{
\footnotesize
\begin{tabular}{ rr| cccccc  } \hline 
&& TP & FP & FN &  Prec & Rec & F0.5 \\ \hline 
 M:PUNCT & A & 189 & 171 & 134 & 0.525 & 0.5851 & 0.536 \\ 
   & B & 203 & 132 & 133 & 0.606 & 0.6042 & 0.6056 \\ 
   & C & 95 & 96 & 80 & 0.4974 & 0.5429 & 0.5059\\ \hdashline
   
R:VERB & A &21&       60 &      113 &     0.2593 &  0.1567   &0.2293 \\
& B & 17&        55&        113&       0.2361&    0.1308&    0.2033 \\
& C & 6&         43&        51&        0.1224&    0.1053&    0.1186 \\ \hline 

\textsc{m} & A & 318 & 436 & 372 &  0.3703 &	0.3571 &	0.1691 \\
& B & 336&347&344& 0.4919&0.4941&	0.2458\\ 
& C & 157&222&168& 0.4142	& 0.4830	&0.2180 \\ \hline


\end{tabular}
}
\caption{Detailed breakdown evaluation results for the  most frequent errors, and \textsc{m}issing operation errors (FT GPT2, zero-shot). 
}\label{label-by-label}
\label{label-by-label-eval}
\end{table}

\paragraph{Is recall higher than precision in prompting GPT for the GEC task?}

Consistent higher recall compared to precision showcases a tendency of over-correction in prompting GPT for the GEC task. We have observed that proficiency levels A and B, however, do not exhibit such a propensity. It holds true even for GPT-3.5, where recall consistently surpasses precision. Nevertheless, the difference between precision and recall measurements in levels A and B is considerably smaller compared to level C.


\paragraph{Results using various F-scores}
Table~\ref{different-fscores} shows results of FT GPT-2 and GPT-3.5 obtained with different F-scores, where $\beta=$ 0.5, 1, and 2.
The result implies that FT GPT-2 is less prone to over-correction in comparison to GPT-3.5 because the F2 scores are mostly higher in GPT-3.5.
In traditional approaches in GEC, such as SOTA results in Table~\ref{gpt2-xl-results}, where the total numbers of TP and FP are relatively small, F0.5 would be relevant to measure GEC results.
Since recent approaches by prompting GPT in the GEC task illustrate much higher numbers, especially FP, it appears that the F1-score proves to be a more effective indicator in GEC results. 

\begin{table}[!ht]
\centering
{\footnotesize
\begin{tabular}{c |ccc | ccc} \hline 
& \multicolumn{3}{c|}{FT GPT-2} & \multicolumn{3}{c}{GPT-3.5} \\
& F0.5 & F1 & F2 & F0.5 & F1 & F2 \\\hline 
A &  0.4192 &0.4032 & 0.3885 & 0.3784 & 0.4030 & 0.4310 \\
B &  0.4210 &0.4010 & 0.3827 & 0.3291 & 0.3625 & 0.4034 \\
C &  0.3310 &0.3388 & 0.3470 & 0.2199 & 0.2680 & 0.3430 \\ \hdashline
all & 0.3907 & 0.4029 & 0.3792 & 0.3590 & 0.3230 & 0.4040 \\ \hline

\end{tabular}
}
\caption{Different F-scores with F0.5, F1 and F2. FT GPT-2 results are based on 0-shot, while GPT-3.5 (\texttt{text-davinci-003}) results are based on 4-shot.}
\label{different-fscores}
\end{table}






\paragraph{Comparison between prompting GPT and SOTA}

State-of-the-art (SOTA) results continue to demonstrate superior performance  compared to prompting GPT in the GEC task in all aspects of results including precision and recall measures regardless of proficiency levels.
Our assumption is primarily based on the fact that SOTA models are usually subjected to extensive fine-tuning processes. 


\paragraph{Discussion}

In this section, we present the evaluation outcomes using our own implementation to count the numbers of TP, FP, and FN, which are different from the ERRANT scores.
We leave it as future work to further investigate and explore potential improvements. 

Additionally, while we examine a correlation between {proficiency level C and native} in prompting GPT in GEC as shown in Table~\ref{c-vs-native}, we are unable to identify any comparable behavior in prompting GPT in GEC for native-like proficiency C and native proficiency.
\citet{hawkins-buttery:2010} analyze that some error types are more notable in B1 and B2 levels than C1 and C2 levels, such as missing preposition and form of determiner. For example, there are more errors like missing preposition (\texttt{M:PREP}) or replacement of determiners (\texttt{R:DET}) in B than in C, which confirm what the previous work proposes. 
Table~\ref{RDET-vs-MPREP} shows a behavior of prompting GPT in the GEC task proficiency specific errors, in which finding their correlation could be excessively challenging because of the performance of GEC for proficiency level C. 
We consider results of the proficiency level C as unnatural behavior, which deviates significantly from what is considered typical prompting GPT in GEC. We also leave it as future work.

\begin{table}[!ht]
    \centering
{\footnotesize
\begin{tabular}{ c ccc ccc} \hline 
& TP & FP  & FN & Prec & Rec & F0.5 \\\hline 
C & 383 & 792 & 703 & 0.326 & 0.3527 & 0.331 \\
N & 2429 & 3474 & 4103 & 0.4115 & 0.3719 & 0.4029\\ \hline 
\end{tabular}
}    \caption{Results between proficiency level C and native}
    \label{c-vs-native}
\end{table}

\begin{table}[!ht]
    \centering
{\footnotesize
\begin{tabular}{rc  | ccc ccc} \hline 
     & & TP & FP & FN & Prec & Rec & F0.5 \\\hline 

M:PREP & B &24&       29&       31&       0.4528&   0.4364&   0.4494 \\
& C & 9 &        23  &      17 &       0.2812 &   0.3462 &   0.2922 \\\hdashline
R:DET & B & 15&        30&        41&        0.3333&    0.2679&    0.3178 \\
& C & 7  &       12  &      23 &       0.3684 &   0.2333 &   0.3302 \\\hline 
\end{tabular}
}    
\caption{Detailed breakdown evaluation results for \texttt{M:PREP} and \texttt{R:DET}}\label{RDET-vs-MPREP}
\end{table}

\section{Conclusion} \label{conclusion-section}
In this paper, we investigated the strengths and limitations of prompting GPT for the GEC task based on different language proficiency levels. 
We used our own implementations to calculate relevant metrics for label-by-label analysis, which are different from the current standard ERRANT scores by using m2 files.
We observed a tendency of over-correction in prompting GPT, and it is more obvious in the recent version of GPTs, where recall consistently surpasses precision. 
Additionally, since prompting GPT generates much higher false positive numbers in results, the F1-score, rather than the F0.5-score, would be a more effective measure in GEC results.

\section{Ethics Statement}
To confirm Behavioural Research Ethics at the University of British Columbia,\footnote{\url{https://ethics.research.ubc.ca/behavioural-research-ethics}} 
authors have obtained a certificate of the Tri-Council Policy Statement: Ethical Conduct for Research Involving Humans (TCPS 2): Course on Research Ethics (CORE-2022).\footnote{\url{https://tcps2core.ca}}

\section*{Acknowledgement}
This work was supported in part by Oracle Cloud credits and related resources provided by Oracle for Research.

\bibliographystyle{plainnat}
\bibliography{references}
\end{document}